\documentclass[letterpaper, 10 pt, conference]{ieeeconf}  

\IEEEoverridecommandlockouts                              

\usepackage{graphics} 
\usepackage{epsfig} 
\usepackage{amsmath} 
\usepackage{amssymb}  
\usepackage{ upgreek }
\usepackage[linesnumbered,ruled]{algorithm2e}
\usepackage[utf8]{inputenc}

\newcommand{\etal}{\MakeLowercase{\textit{et al.\ }}}
\usepackage{xcolor}

\usepackage{graphicx}
\usepackage{float}
\usepackage{amsmath}

\usepackage{cuted}

\usepackage{lipsum,amsmath,multicol}
\usepackage[official]{eurosym}

\usepackage{subfigure}
\usepackage{algpseudocode}
\usepackage{epstopdf}
\usepackage{lipsum}
\usepackage{tabularx}
\usepackage{multicol}
\usepackage{multirow}
\usepackage[leftcaption]{sidecap}
\usepackage{booktabs}

\usepackage{amssymb}
\usepackage{arydshln}

\usepackage[hidelinks]{hyperref}
\usepackage{paralist}

\usepackage[nolist]{acronym} 
\newacro{fem}[FEM]{finite element modeling}
\newacro{fea}[FEA]{finite element analysis}
\newacro{pneunets}[Pneu-nets]{pneumatic networks}
\newacro{3d}[3D]{ three-dimensional}
\newacro{cad}[CAD]{computer-aided design}
\newacro{gui}[GUI]{Graphical User Interface}
\newacro{hbsf}[HBSF]{hybrid bending soft finger}
\newacro{hbsa}[HBSA]{hybrid bending soft actuator}
\newacro{spa}[SPA]{soft pneumatic actuator}
\newacro{sma}[SMA]{shape memory alloy}
\newacro{rom}[ROM]{Ranges of Motion}

\title{\LARGE \bf
Biomimetic Evaluation of an Underwater Soft Hand Through\\
Deep Learning-based 3D Pose Reconstruction}

\author{Haihang~Wang,
        He~Xu, 
        and~Yihan~Meng

\thanks{This work was supported by the Natural Science Foundation of China under Grant 51875113, Natural Science Joint Guidance Foundation of the Heilongjiang Province of China under Grant LH2019E027. The first author thanks China Scholarship Council (CSC) for financial support through No. 201906680045. (Corresponding author:~He~Xu)}
\thanks{
The authors are with College of Mechanical and Electrical Engineering, Harbin Engineering University, China. (e-mail: wanghaihang@hrbeu.edu.cn; railway\_dragon@sohu.com; myh0430@hrbeu.edu.cn)}
}

\begin{document}
	
	\maketitle
	\thispagestyle{plain}
	\pagestyle{plain}
	
	\begin{abstract}
	Soft robotic hand shows considerable promise for various grasping applications. However, the sensing and reconstruction of the robot pose will cause limitation during the design and fabrication. In this work, we present a novel 3D pose reconstruction approach to analyze the grasping motion of a bidirectional soft robotic hand using experiment videos. The images from top, front, back, left, right view were collected using an one-camera-multiple-mirror imaging device. The coordinate and orientation information of soft fingers are detected based on deep learning methods. Faster RCNN model is used to detect the position of fingertips, while U-Net model is applied to calculate the side boundary of the fingers. Based on the kinematics, the corresponding coordinate and orientation databases are established. The 3D pose reconstructed result presents a satisfactory performance and  good  accuracy. Using efficacy coefficient method, the finger contribution of the bending angle and distance between fingers of  soft robot hand is analyzed by compared with that of human hand. The results show that the soft robot hand perform a human-like motion  in both single-direction and bidirectional grasping.
	\end{abstract}

\section{Introduction}

Soft robots have advantages over rigid robots due to their flexibility, compliance, and adaptability to the object and environment \cite{icsaah3690}. Dexterity of the human hand is of tremendous importance for exploring and interacting with the world and a top design objective for robotic hands \cite{6298887}. Fluidic elastomer actuators provide particular advantages over rigid underwater actuators, owing to their flexibility, light weight, low cost, and waterproofing features inherently endowed by soft materials  \cite{Shintake30,SUZUMORI1996135}. Bidirectional actuators that are capable of both inward and outward bending show potential on improving the grasping capability of grippers. Soft robotic hands with anthropomorphic design are emerging and being developed by researchers. Compared to joint-like bending motion with a small bend radius and a large bend range of human fingers, soft bending actuators lack the capability of controlling the angle of the fingertip. An opposable and flexible palm enables fingers with more grasping dexterity.

Soft robotic hands with anthropomorphic design are emerging and being of interest of many scientists.
Examples of the soft biomimetic hands include RBO Hand 2 \cite{RN1755} and 3 \cite{RBOHand33}, the soft neuroprosthetic hand \cite{RN1763}, BCL-26 Anthropomorphic hand  \cite{RN1763}, human-inspired soft hand with active palm \cite{WHH_RAM} and so on. Although various SRHs have been developed for safe, adaptable grasping, most of them still lack sensory feedback \cite{8821518}. For achieving the proprioception in soft robots, a large variety of flexible tactile sensors provides a much convenient way for the soft bending actuators. Most of these sensors are based on capacitive, resistive and piezoresistive, or
optical principles \cite{s16122001, 7547359,8594270}. 
However, the capacitance of capacitive sensors is affected by the change of the gap between the conductive plates on the imposition of external force. Resistive sensors generate error in electrical resistance when mechanical deformation occurs during working. Optical fiber sensors are immune to magnetic fields and are inherently safe for interacting with human beings \cite{kappassov}. Besides, the addition of the sensors into the soft material will greatly increase the risk of  rupture under higher pressure. Moreover, integrating the sensor into soft actuator is usually cumbersome during fabrication. 

Compared to these sensing modalities, directly acquiring the motion information through experiment video has no influence on the design and fabrication. 
Ref. \cite{Digumarti105}  reconstructed numerically soft deformable morphologies based on experiment images. Inoue \etal \cite{Inoue110} skeletonized the dynamics of soft robot body from video. Using the image from three cameras, Ref. \cite{7139340} reconstructed the swimming motion of the underwater soft robotic octopus.
However,  these vision-based reconstruction method is aiming at process arm-like soft actuators, which is unsuitable for the complex motion of soft robotic hands,  especially when the robotic hand grasps with object. 

\begin{figure}[!tbp]
    \centering
    \includegraphics[width= 85 mm]{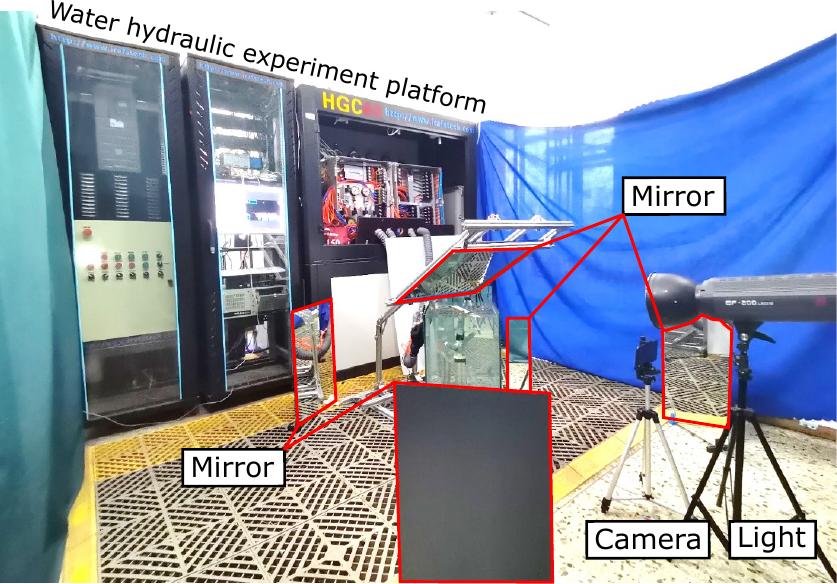}
    \caption{The 3D imaging system for bidirectional grasp.}
    \label{IROSFigMirrorPicture}
\end{figure}

In this paper, we present a novel 3D reconstruction approach  to analyze the complex grasping motions of soft robots in underwater environment through experiment videos. The videos were captured by the 3D imaging system, as shown in Fig. \ref{IROSFigMirrorPicture}. The proposed reconstruction method perform the reconstructed hand pose when the hand is both with and without grasp objects.

\section{3D Imaging System}
\subsection{Experimental Apparatus}
The soft robotic hand used in this study was based on our previous work \cite{WHH_RAL}, which works in underwater environment and is able to achieve bidirectional grasping.
During bidirectional grasping, the image from single view can not fully present and collect comprehensive hand grasp information. Thus, we developed a 3D imaging system to providing all-sided images about this novel bidirectional grasp.
As shown in Fig. \ref{IROSFigMirrorPicture}, the 3D imaging system consists of three parts. One is the water hydraulic experiment platform, which powers, sense and control the water pressure into the soft hand. The second includes the soft robotic hand, water tank and metal support. The third is the one-camera-multiple-mirror imaging device.

\subsection{One-Camera-Multiple-Mirror Imaging Device}
\begin{figure}[tbp]
    \centering
    \includegraphics[width= 75 mm]{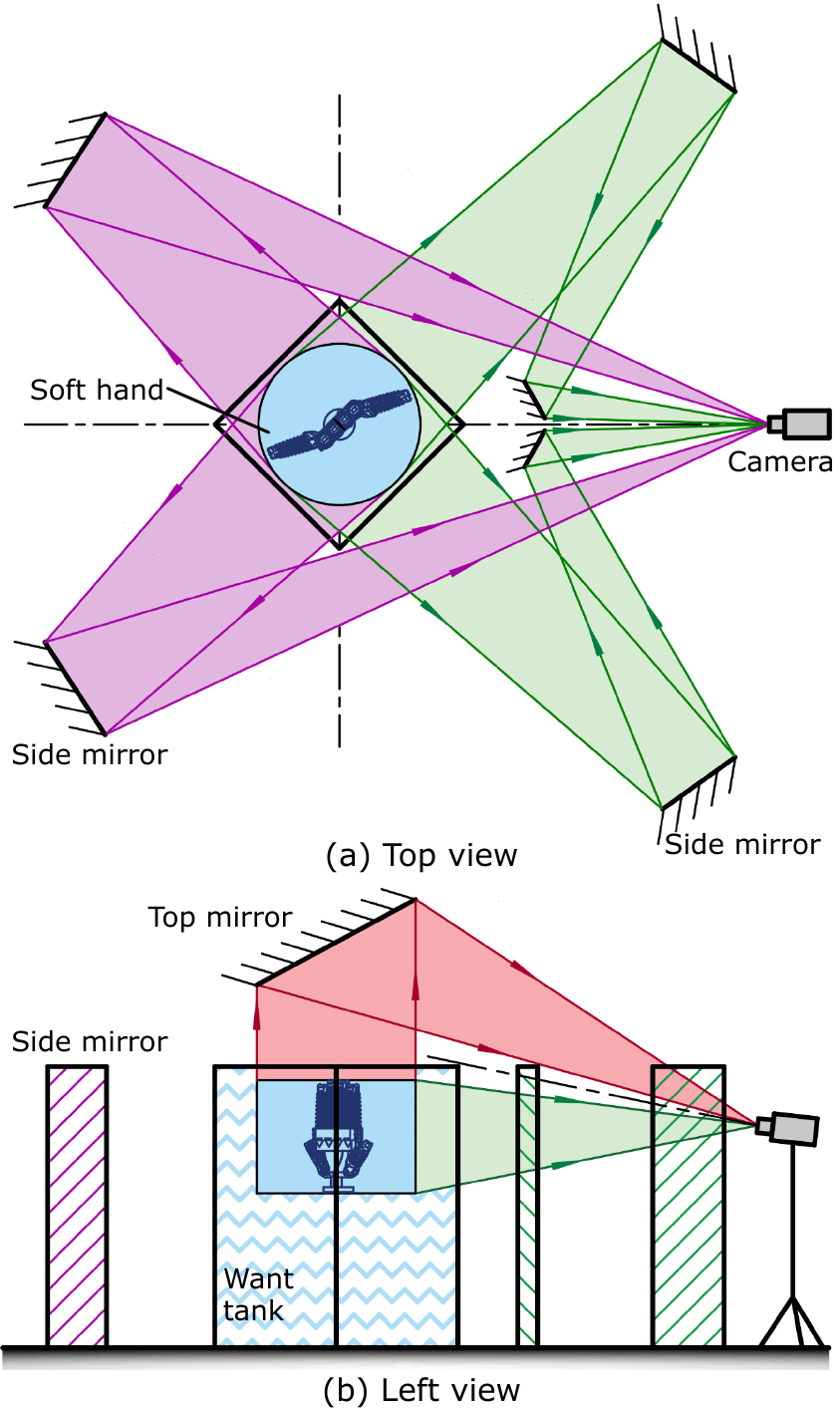}
    \caption{The principle drawing of the light paths from the top, front, back, left and right directions.}
    \label{IROSFigMirrorDrawing}
\end{figure}

The one-camera-multiple-mirror imaging device consists of 1 camera and 7 mirrors placed around the observed water tank. Fig. \ref{IROSFigMirrorDrawing} describes its imaging principle. The image information of the soft hand will be reflected by the mirrors and captured by one camera. As shown in Fig. \ref{IROSFigMirrorDrawing}(a), the arrangement of the camera and mirrors has been optimized. To increase the resolution of the experimental images, and reduce the imaging distance, the light path was designed as close to the tank as possible. To decrease the effect of camera distortion, the camera was arranged up to balance the top and horizontal directions.

\section{3D Reconstruction Approach}

\subsection{Deep Learning Based Feature Detection}
\begin{figure*}[!tbp]
    \centering
    \includegraphics[width= 175 mm]{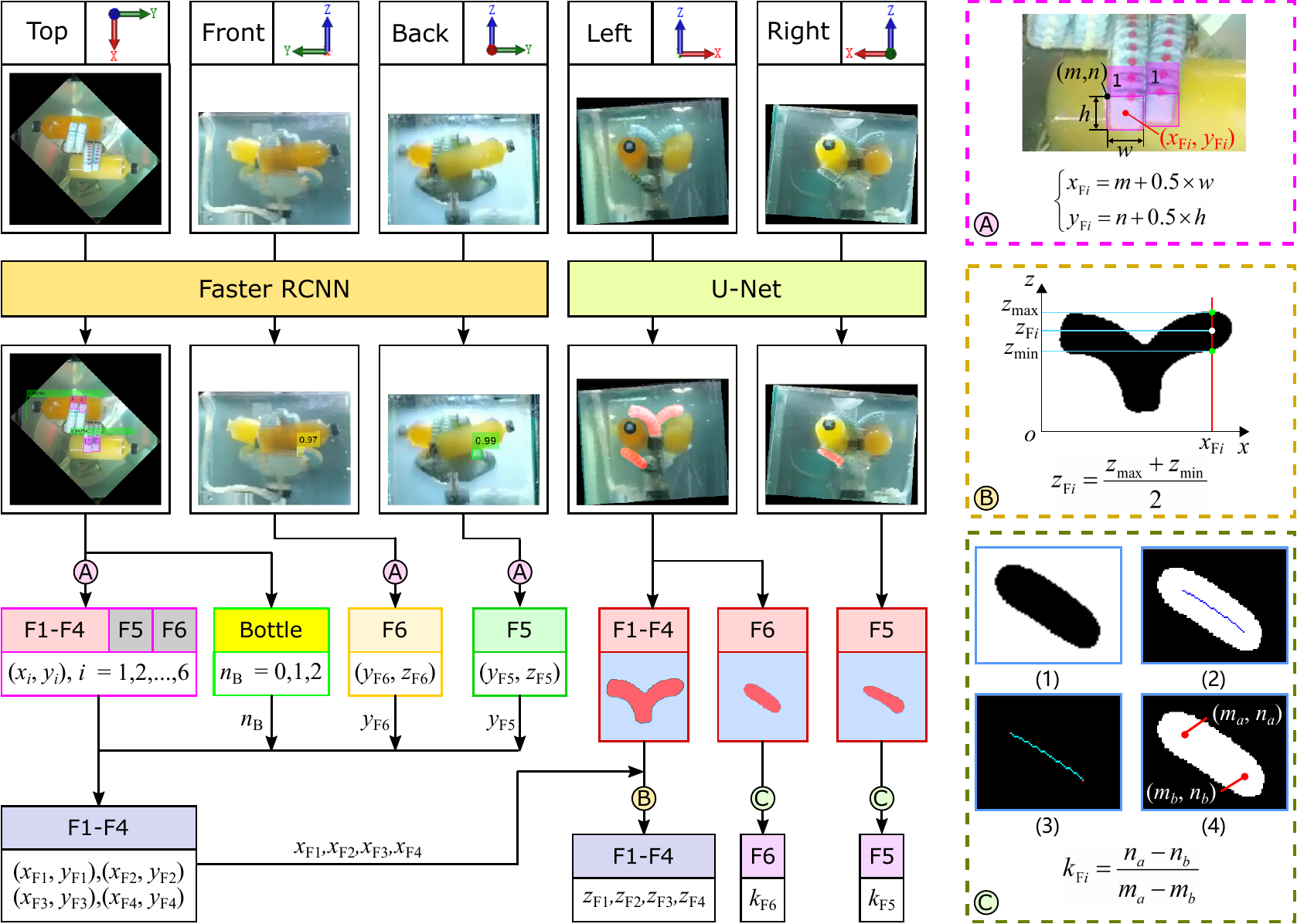}
    \caption{Finger feature detection framework based on deep learning. (A): Calculate the coordinate of the center point in the top image; (B): Locate the center point of fingertip according to $x_{\rm{F}i}$ in the left image; (C): Calculate the slope of thumb in the left/right images based on the skeletonization image processing;
    (F1: Index finger; F2: Middle finger; F3: Ring finger; F4: Little finger; F5: Main thumb; F6: Vice thumb.).}
    \label{IROSFigDeeplearningMappingFrameworkV5}
\end{figure*}

Deep learning methods, particularly artificial neural networks, provide a powerful tool to learn the motion feature of soft robots \cite{9426391}. With the 5 experiment videos from top/front/back/left/right views, a finger feature detection framework has been developed for achieving the position and orientation of the 6 soft fingers.  In this study, we always define F1 represents the index finger, F2 represents the middle finger, F3 represents the ring finger, F4 represents the little finger and F5/F6 represents the main/vice thumb. 
As shown in Fig. \ref{IROSFigDeeplearningMappingFrameworkV5}, two deep learning algorithm, Faster RCNN and U-Net, are applied to acquire the 3D pose data of the soft hand. 
The key data of the 3D reconstruction model generated in the figure includes the current frame number, the x, y, and z coordinates of the 6 orange fingertips, the finger bending angle, the finger deflection angle and the palm bending angle. These 7 key data are recorded.

The  identification of fingertip and grasp object was based on the Faster R-CNN method developed by Ren \etal. \cite{NIPS201514bfa6bb}, which is a mainstream deep learning method in object detection. The average number of the training samples for the fingertips in the top, front and left view and the bottle is 93. We got a stable and satisfactory accuracy after finishing an average 1000 iterations.  
The finger side-boundary was based on the U-Net method developed by Ronneberger \etal\cite{Ronneberger8}, which provides an effective network and training strategy for image segmentation. We used 109 training samples and implemented 10 epochs.

Using the Faster RCNN and U-Net models, the $(x, y,z)$ coordinate information of the index, middle, ring and little fingers are calculated and collected. Note that the pixel position data transmitted between different view images has been calibrated by pixel/length scale. For the main and vice thumbs, we get the $(y,z)$ coordinate information and their orientation in the left view.

\subsection{Kinematic Modeling}
\begin{figure}[!tbp]
    \centering
    \includegraphics[width= 75 mm]{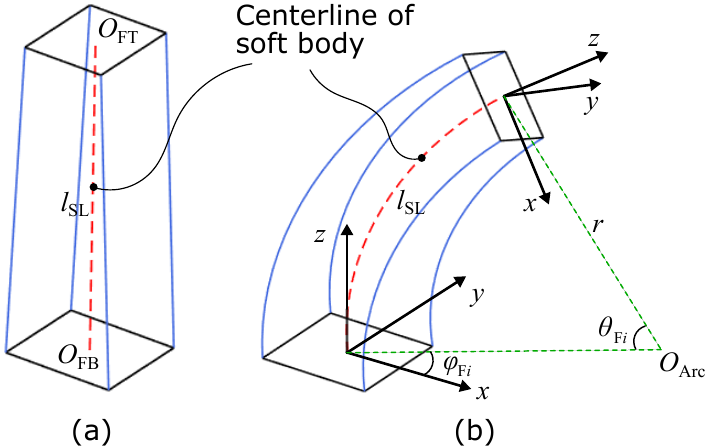}
    \caption{Kinematics modeling of the soft actuator. (a) The soft actuator at linear state. (b) The soft actuator at bending state.}
    \label{IROSFigPCCmodeling}
\end{figure}

The robotic hand  kinematics is described using homogeneous transformation matrix, which maps the bending angles of the soft actuators to the major parameters of the robot configuration space. 
As shown in Fig. \ref{IROSFigPCCmodeling}, the position and  pose of the soft finger can be quantified by the points $O_{\rm{FB}}$ and $O_{\rm{FT}}$ on the centerline. As illustrated by the red dashed line, the length of the strain limiting layer $l_{SL}$ is constant.  The curvature and place can be calculated by the
trigonometric relationship \cite{8722773}. The origin is at the base part of the soft hand. The geometrical relations of the soft hand provides a means of determining the relative position of key points on the outlines.

For the bending segment of soft finger, the orientation angle ${\phi _{{\rm{F}}i}}$ represents the rotation around the $z$-axis, curvature angle ${\theta _{{\rm{F}}i}}$ represents the rotation around the $y$-axis, where $i$ indicates the $i$-th finger. The transformation matrix for the flexible segment of soft finger is expressed as Eq. \ref{EqTFIRZ}. That of the soft palm can be acquired by a similar, simplified transmission.

\begin{equation}\label{EqTFIRZ}
\begin{aligned}
{{\bf{T}}_{{\rm{Fi}}}} =& \left[ {\begin{array}{*{20}{c}}
{{{\bf{R}}_z}({\phi _{{\rm{F}}i}})}&0\\
0&1
\end{array}} \right] \cdot \left[ {\begin{array}{*{20}{c}}
1&{{{\bf{p}}_{{\rm{F}}i}}}\\
0&1
\end{array}} \right]\\
 \cdot &\left[ {\begin{array}{*{20}{c}}
{{{\bf{R}}_y}({\theta _{{\rm{F}}i}})}&0\\
0&1
\end{array}} \right] \cdot \left[ {\begin{array}{*{20}{c}}
{{{\bf{R}}_z}( - {\phi _{{\rm{F}}i}})}&0\\
0&1
\end{array}} \right]
\end{aligned}
\end{equation}

\[{{\bf{p}}_{{\rm{F}}i}} = {[\begin{array}{*{20}{c}}
{{r_{{\rm{F}}i}}(1 - \cos {\theta _{{\rm{F}}i}})}&0&{{r_{{\rm{F}}i}}\sin {\theta _{{\rm{F}}i}}}
\end{array}]^{\rm{T}}}\]

\begin{figure}[tbp]
    \centering
    \includegraphics[width= 85 mm]{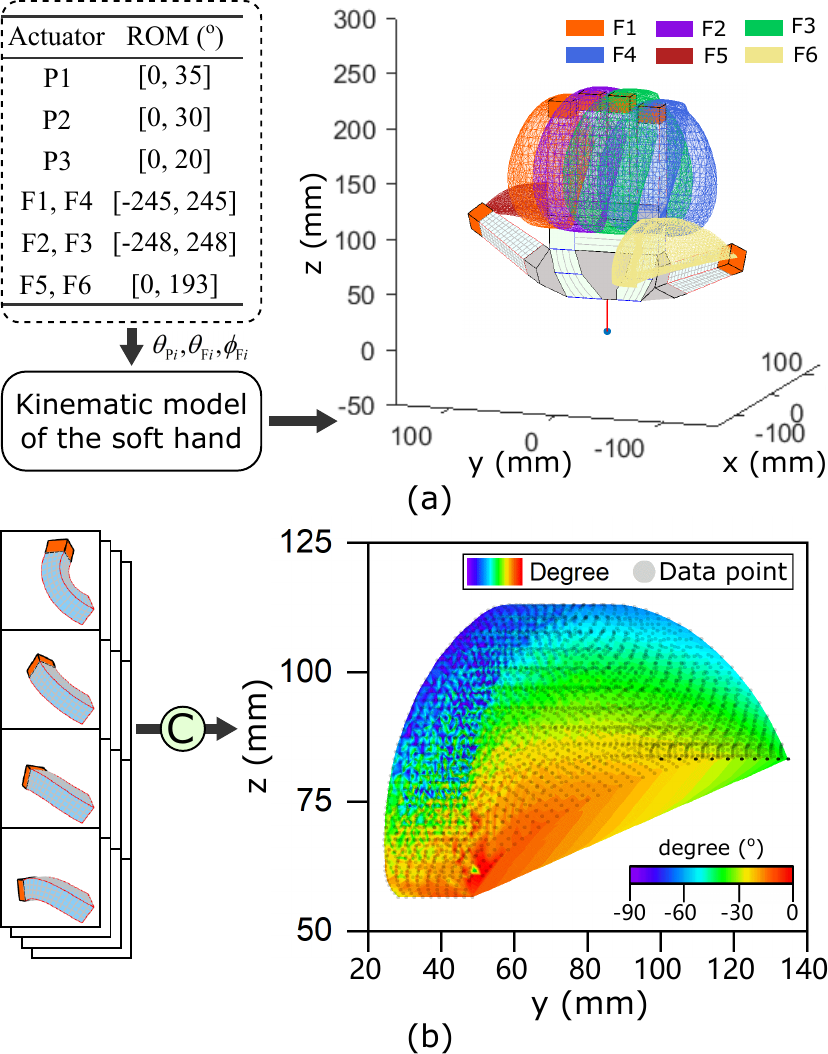}
    \caption{The establishment of (a) the  workspace database and (b) the slope angle database.}
    \label{IROSFigWorkspaceKCountorV3}
\end{figure}

 The soft hand configuration and pose can be simulated by importing the angle variables ${\phi _{{\rm{P}}j}}$, ${\phi _{{\rm{F}}i}}$ and ${\theta _{{\rm{F}}i}}$.  ${\phi _{{\rm{P}}j}}$ represents the bending angle of the $j$-th soft palm actuator.  Workspace sampling was performed in terms of finger level, by selecting values of configuration parameters as ${\phi _{{\rm{P}}j}}$, ${\phi _{{\rm{F}}i}}$, ${\theta _{{\rm{F}}i}}$ within certain range, as shown in Fig. \ref{IROSFigWorkspaceKCountorV3}(a). The sampled data points for each parameter were spaced by the same value. It is intuitive that the denser the sampling points is, the higher the accuracy should be. However, larger the database size is, slower the mapping process would be.
 Considering the error caused by the feature detection accuracy and to shorten the further process of data mapping, the workspace database including 4000 points was established.
 
 \begin{equation}
    \alpha_{{\rm{F}}i} = \arctan k _{{\rm{F}}i}
\end{equation}

The slope angle  $\alpha _{{\rm{F}}i}$ database was created to help map the side orientation of the thumbs .  As shown in Fig. \ref{IROSFigWorkspaceKCountorV3}(b), we separately simulate the configuration of the main (F5) and vice (F6) thumbs. Following the same image processing method (C), we calculate  the slope angle of all the points in the workspace database from the left/right views.

\subsection{Workspace Mapping Strategies}
\begin{figure*}[!tbp]
    \centering
    \includegraphics[width= 175 mm]{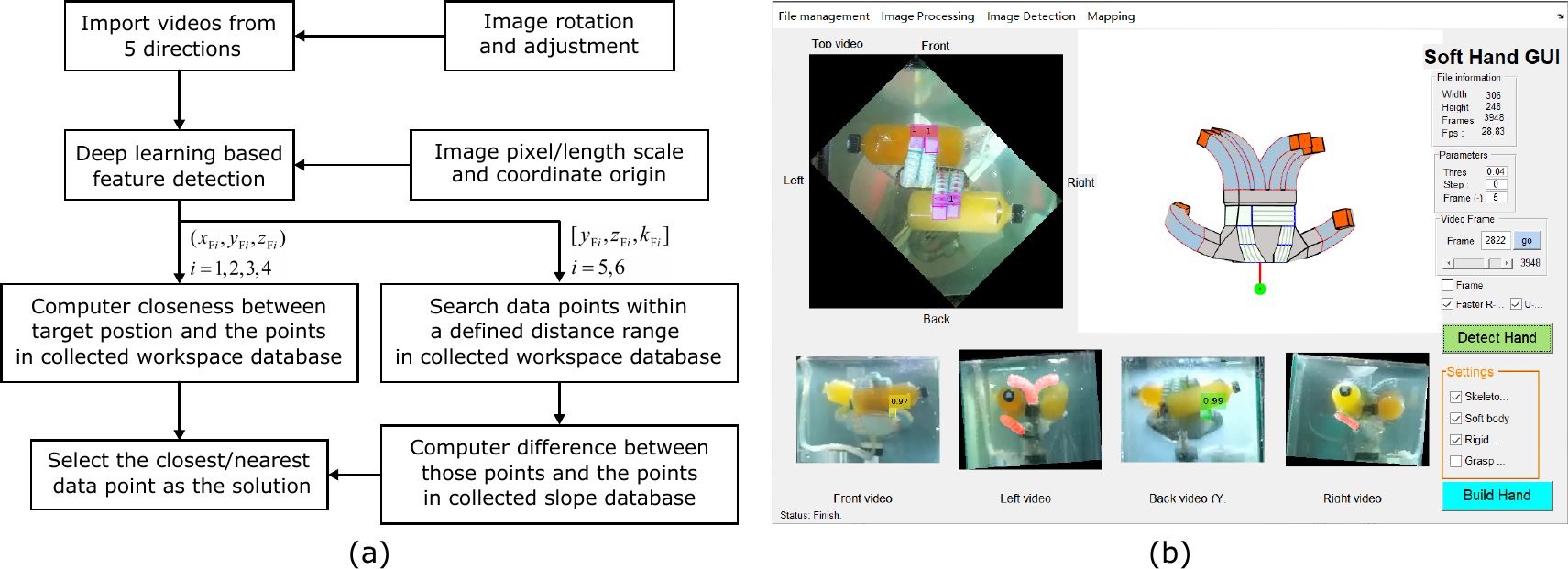}
    \caption{(a) Workspace mapping  approach. (b) The Graphical User Interface (GUI) consisting of image presentation and processing, feature detection,  workspace mapping and 3D pose reconstruction.}
    \label{IROSFigMatlab3DreconstructionFramework}
\end{figure*}

Different  strategies are designed for the workspace mapping of the F1-F4 and F5/F6. As  illustrated in Fig. \ref{IROSFigMatlab3DreconstructionFramework}(a), after achieving the feature detection and coordinate transmission, the searching strategy for F1-F4 is quite straightforward by sequentially comparing closeness in space between each data point and the target position. With respect to that of F5/F6, we first search the data points within a distance range in $y$-$z$ plane. Then, we select the data point that has closet slope angle difference. Finally, the ${\phi _{{\rm{P}}j}}$, ${\phi _{{\rm{F}}i}}$, ${\theta _{{\rm{F}}i}}$ of the mapped data points will be imported in the kinematic model. The 3D pose reconstruction method of the soft hand is integrated in the dedicated program with a GUI, as shown in Fig. \ref{IROSFigMatlab3DreconstructionFramework}(b). It is built on a laptop with MATLAB 2016a and Python 3.7 or higher version on the Windows operating system.

Fig. \ref{IROSFigMatlab3DreconstructionFramework} presents the reconstruction results of the key frames during the bidirectional grasp, which are origin pose single-direction grasping pose, bidirectional grasping   pose and its  grasping pose when with one or two objects (bottle). Comparing the reconstructed pose with the experiment images from the top, front and left view, it is validated that 
the effectiveness and  accuracy of our proposed 3D reconstruction approach is convincing and satisfactory. 

\begin{figure}[!tbp]
    \centering
    \includegraphics[width= 85 mm]{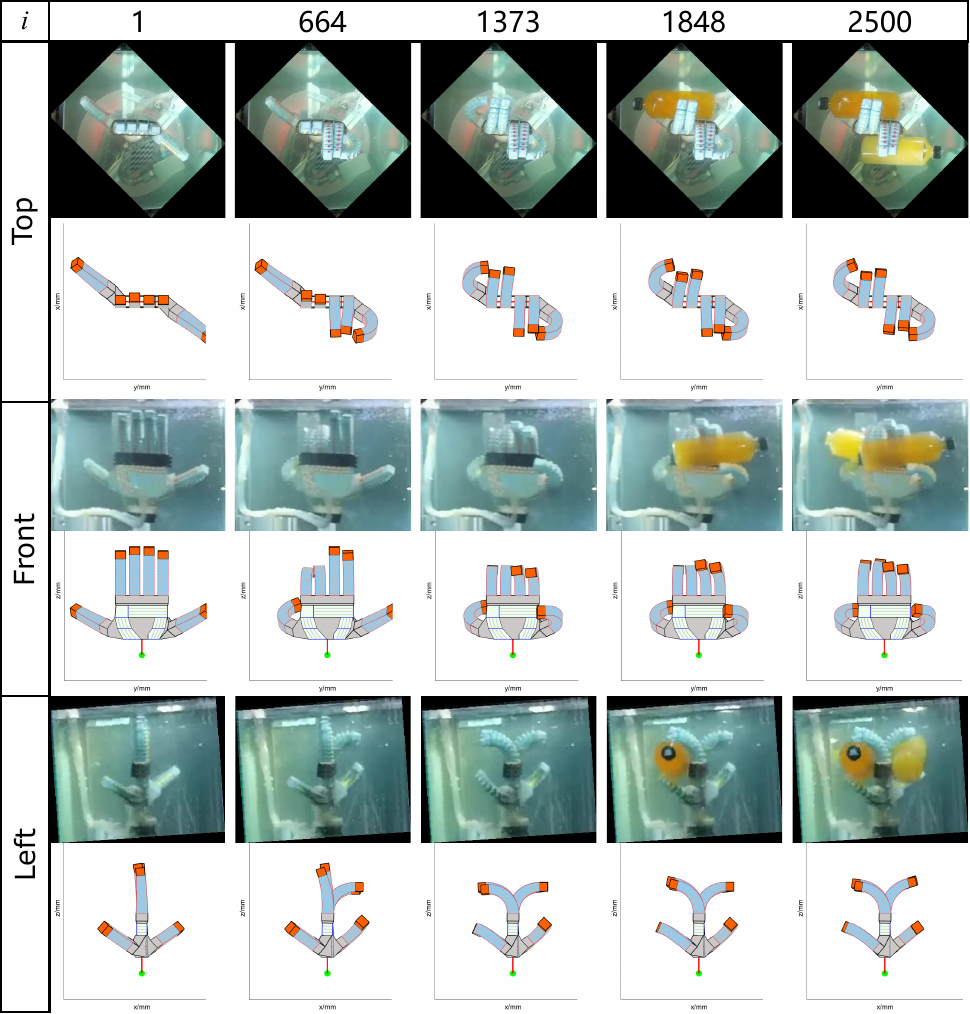}
    \caption{Selected 3D pose reconstruction results of the soft hand during bidirectional grasping.}
    \label{IROSFigMatlab3DreconstructionResult}
\end{figure}

\section{Biomimetic Evaluation and Analysis}

\subsection{Grasp Evaluation}

\begin{figure}[tbp]
    \centering
    \includegraphics[width= 85 mm]{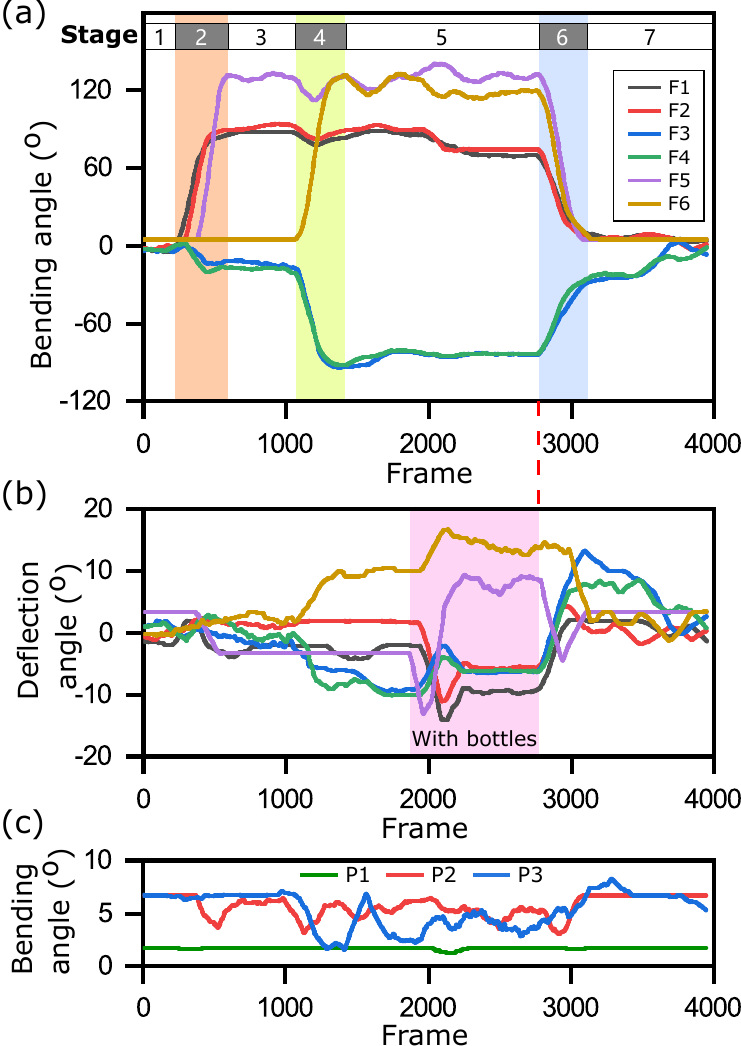}
    \caption{The three key angles of the reconstructed result based on the bidirectional grasping video. (a) The bending angle of soft fingers in the main grasping axis. (b) The deflection angle of soft fingers in the side-deflection axis. (c) The bending angle of soft palm actuators. (P1: Palm actuator 1; P2: Palm actuator 2; P3: Palm actuator 3.)}
    \label{IROSFigSmoothedResult}
\end{figure}

Using the 3D reconstructed algorithm, the experimental video about bidirectional grasping was calculated frame by frame. We skipped the image  in the frames when the researcher is operating the bottle to be grasped, or the bottle is greatly shelter the detection features. To eliminate the data fluctuation, the data in Fig. \ref{IROSFigSmoothedResult} was smoothed using Savizky-Golay method with 1 polynomial order. 
The number of experimental frames is the number of frames in the experimental video. I captured a total of 3948 frames of the experimental video, of which 1 to 3200 frames are the entire capture process, and the experiments from 3200 to 3948 frames are useless and static. Among the 3200 frames, 2758 frames were reconstructed in 3D through Matlab. After manually removing abnormal and obviously incorrect data, a total of 2560 frames were the final valid results, and the algorithm recognition and reconstruction rate was about 80\%.

Human hand dexterity has provided numerous insights on development of robotic hands.
Min \etal evaluated the robot hand with human hand with respect to human-like grasping features \cite{9345451}. Since we aim to using the comprehensive efficacy coefficient to analyze the contribution of each fingers of human/robot hand for successful grasp, we perform experiments on the same grasping motion as single-direction grasping to acquire the motion data of human hand. The 3D hand pose data of human hand was estimated using the pretrained network model developed in Ref. \cite{Ge8953612}. 
Each finger has a total of 4 feature points, plus a common base point of the palm and wrist. A total of 21 feature points. The bending angle of each finger can be obtained from the included angle of the line segment composed of 4 feature points.

\begin{figure}[tbp]
    \centering
    \includegraphics[width= 75 mm]{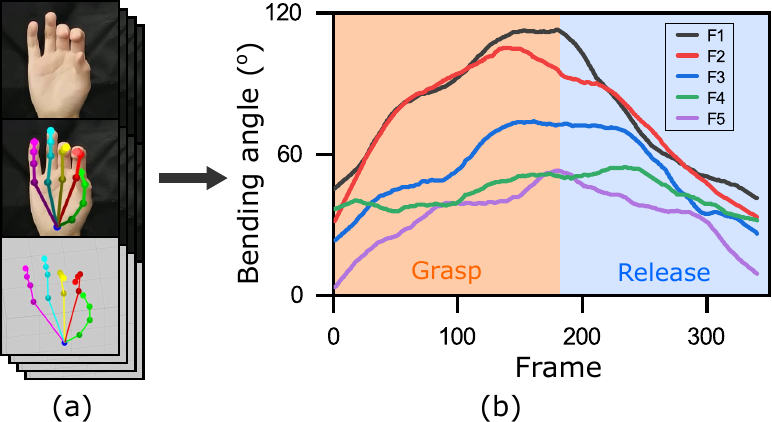}
    \caption{The grasping data of human hand was collected according to the same grasping motion of single-direction grasp. (a) The 3D pose was estimated by human hand grasp image. (b) The bending angle of human fingers by frame.}
    \label{IROSFigHumanHandDetectCVPR}
\end{figure}

\begin{strip}
\begin{align}
Q_{{\rm{S}}i}^{} = {}^{(\sum\limits_{m = 1}^M {w_{{\rm{F}}m}^{}}  + \sum\limits_{n = 1}^N {w_{{\rm{R}}n}^{}} )}\sqrt {\prod\limits_{m = 1}^M {q_{{\rm{S}}i,{\rm{F}}m}^{w_{{\rm{F}}m}^{}}}  \times \prod\limits_{n = 1}^N {{{\left( {q_{{\rm{S}}i,{\rm{R}}n}^{} - s(q_{{\rm{S}}i,{\rm{R}}n}^{} - 1)} \right)}^{w_{{\rm{R}}n}^{}}}} }  - 1\\
Q_{{\rm{B}}i}^{} = {}^{(\sum\limits_{m = 1}^M {w_{{\rm{F}}m}^{}}  + \sum\limits_{n = 1}^N {w_{{\rm{R}}n}^{}} )}\sqrt {\prod\limits_{m = 1}^M {{{\left( {q_{{\rm{B}}i,{\rm{F}}m}^{} - s(q_{{\rm{B}}i,{\rm{F}}m}^{} - 1)} \right)}^{w_{{\rm{F}}m}^{}}}}  \times \prod\limits_{n = 1}^N {q_{{\rm{B}}i,{\rm{R}}n}^{w_{{\rm{R}}n}^{}}} }  - 1
\end{align}
\end{strip}

\begin{figure}[tbp]
    \centering
    \includegraphics[width= 85 mm]{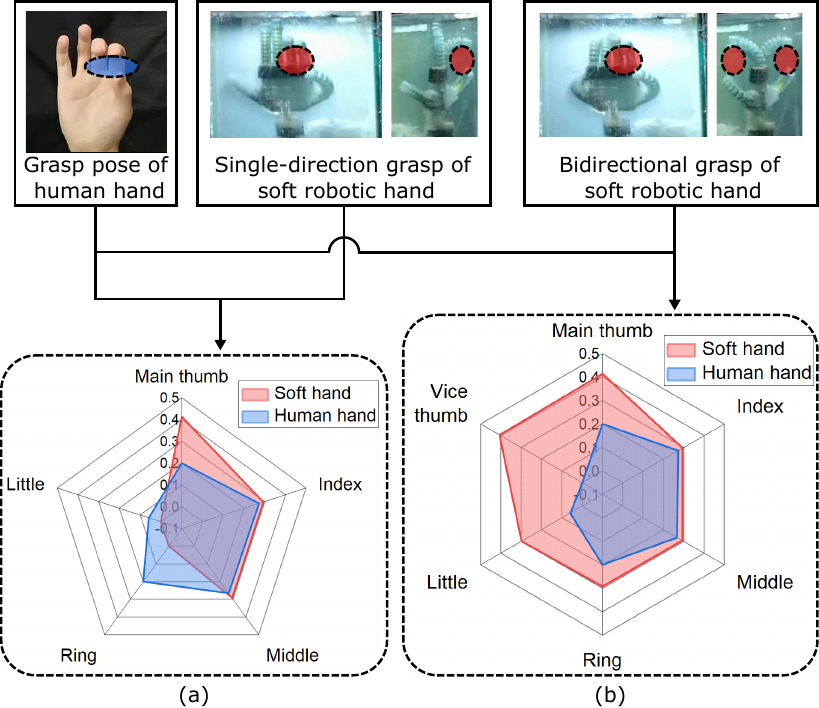}
    \caption{The finger contribution comparison of the bending angle for human and robot hands. (a) Single-direction grasp; (b) Bidirectional grasp.}
    \label{IROSFigBiomimeticRatioComparsion}
\end{figure}

Based on the collected data of the fingers of the soft robot and human hands, the proportion of the bending angle were calculated to analyze the contribution of each fingers. As shown in Figure \ref{IROSFigBiomimeticRatioComparsion}, for the single-direction grasp, the thumb, index and middle fingers act as the main roles for the grasp pose. Beside, the relative high weight of the human ring finger is caused by the 
coupling motion relationship between middle and ring fingers. With respect to the bidirectional grasp, the soft hand performs a prominent advantage compared with human hand. It indicates that the bidirectional grasp fully releases the potential abilities of the ring and little fingers, while promises the quality of the single-direction grasp.

In the grasping work of software manipulators or human hands, it is necessary to conduct coupled evaluation of multi-finger and complex data from multiple data sources. Considering the data of one finger alone, it is impossible to evaluate its contribution to the final grasping result as a whole. The efficacy coefficient method can reflect the multi-factor connection of the bionic efficacy index in the multi-finger coupled bionic grasping process, and can quantify the comprehensive system, which is convenient for comparison. Using Pearson correlation analysis and sliding window dynamic correlation coefficient analysis, the bionic correlation between human hand and soft hand was compared and analyzed. In view of the characteristic bionic two-way grasping of the underwater soft hand, based on the deep learning software for 3D reconstruction of the soft hand, the dynamic data of the one-way grasping by the human hand is collected at the same time, and a bionic grasping evaluation system based on the efficacy coefficient method is established, and the dynamic bionic correlation is carried out. Analysis, the correlation analysis of the bending angles of the fingers of the soft hand and the fingers of the human hand when completing the one-way or two-way grasping task. Considering the contribution of the bending angle of each finger and the grasping space of each fingertip to the grasping success, and converting it into grasping efficacy, the dynamic bionic correlation analysis is carried out on the process of grasping and releasing the target object.

\begin{figure}[tbp]
    \centering
    \includegraphics[width= 85 mm]{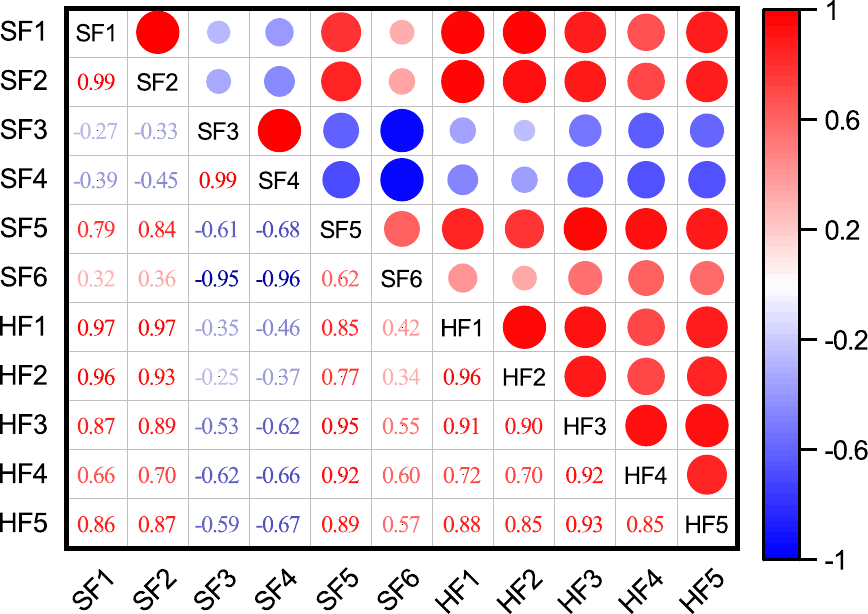}
    \caption{The biomimetic correlation analysis of the fingers of human/soft hands. (SF: Soft finger; HF: Human finger).}
    \label{IROSFigBiomimeticCorelation}
\end{figure}

\subsection{Comprehensive  Efficacy Evaluation}
The efficacy coefficient method is a method mostly used for comprehensive evaluation of economic benefits. In the grasping work of software manipulators or human hands, it is necessary to conduct coupled evaluation of multi-finger and complex multi-data source data. Considering the data of a finger alone, it is not possible to evaluate it as a whole. Contribution to the final crawl result. The efficacy coefficient method can reflect the multi-factor relationship of the bionic efficacy index in the process of multi-finger coupled bionic grasping, and can calculate the comprehensive system quantitatively, which is convenient for comparison.

A successful grasp is the collaboration result of multiple fingers. During the grasping motion of soft/human hands, the complex motion parameters generated by the fingers needs to be comprehensively analyzed. Efficacy coefficient method, which evaluates the comprehensive efficacy of the studied object, is used to evaluate the  grasping efficacy.

Let ${x_{ki}}$ ($k = 1,2, \cdots ,K$) present the values of motion indexes at $i$-th frame. $k$ is the number of evaluation indexes. In general, the coefficient value of $q_{ki}$ is between the objective and the worst values, that is, $x_{ki}^{{\rm{(s)}}} \le {x_{ki}} \le x_{ki}^{{\rm{(h)}}}$. A single efficacy coefficient, namely ${q_{ki}}$, can be described by the following function:

\begin{equation}\label{Eqqkifi}
    {q_{ki}} = {f_i}({x_{ki}}) = \frac{{{x_{ki}} - x_k^{{\rm{(s)}}}}}{{x_k^{{\rm{(h)}}} - x_k^{{\rm{(s)}}}}} + 1
\end{equation}

\noindent where $x_{ki}^{{\rm{(h)}}}$ denotes the satisfactory value, which usually  the optimal limit of the grasping task, and $x_{ki}^{{\rm{(s)}}}$ denotes the impermissible value, which is usually the most unfavorable limit for the task. Then, we establish the  total efficacy coefficient:

\begin{equation}\label{EqQIW12K}
    {Q_i} = {}^{2 \times ({w_1} + {w_2} +  \cdots  + {w_K})}\sqrt {q_{1i}^{{w_1}} \times q_{2i}^{{w_2}} \times  \cdots  \times q_{Ki}^{{w_K}}}
\end{equation}

\noindent where $w_k$ represents the weight of $k$-th evaluation index. $w_k$ can be acquired by Eq. \ref{EqWKXKIH}.

\begin{equation}\label{EqWKXKIH}
    w_k^{} = \frac{{x_{ki}^{{\rm{(h)}}} - x_k^{{\rm{(s)}}}}}{{\sum\limits_{k = 1}^{K} {\left( {x_k^{{\rm{(h)}}} - x_k^{{\rm{(s)}}}} \right)} }} 
\end{equation}

For the bidirectional grasping motion in this study, the  total  efficacy coefficient $Q_i$ are the combined index that expresses one or two successful object grasping. Thus, $Q_i$ can be expressed as the sum of two parts:

\begin{equation}\label{EqQIQSI}
    {Q_i} = {Q_{{\rm{S}}i}} + {Q_{{\rm{B}}i}}
\end{equation}

\noindent where $Q_{{\rm{S}}i}$ denotes the efficacy coefficient for single-direction grasping and $Q_{{\rm{B}}i}$ denotes the extra efficacy coefficient for the  bidirectional grasping.

${q_{{\rm{S}}i,{\rm{F}}m}^{}}$ and ${q_{{\rm{B}}i,{\rm{F}}m}^{}}$  are the variation indexes that contributing to the front-directional grasp during the single-direction and bidirectional grasping status. $M$ is the number of front-directional grasping indexes. 
${q_{{\rm{S}}i,{\rm{R}}n}^{}}$ and ${q_{{\rm{B}}i,{\rm{R}}n}^{}}$  are those of the reversed-directional grasp. And $N$ is the number of reverse-directional grasping indexes.  When in single-direction grasping status, $s=0$. When in bidirectional grasping status, $s=1$.

\subsection{Biomimetic Analysis}

\begin{figure}[tbp]
    \centering
    \includegraphics[width= 85 mm]{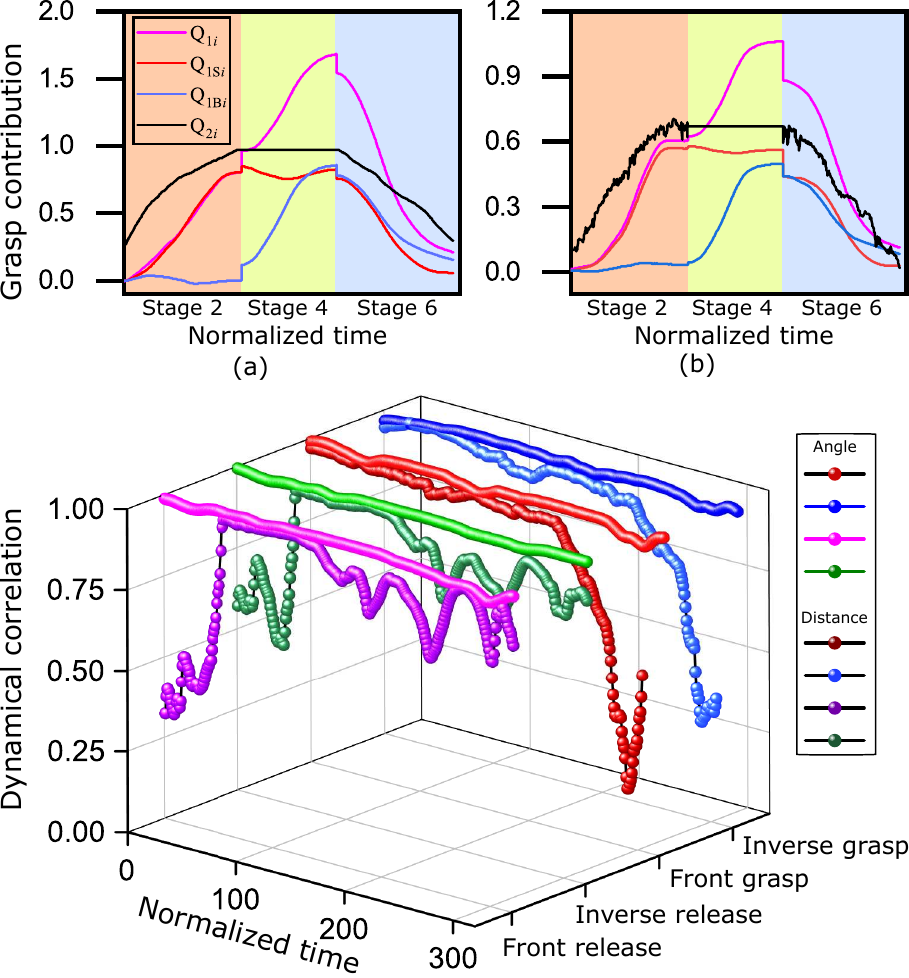}
    \caption{The biomimetic evaluation by comparing the soft robotic hand with human hand during the single-direction grasp, the bidirectional grasp and grasp release. (a) The grasp contribution of fingers about bending angle. (b) The grasp contribution of fingers about fingertip distance.}
    \label{IROSFigBiomimetic}
\end{figure}

The  efficacy coefficients of both human and robot hands $Q_{{1}i}^{}$, $Q_{{1\rm{F}}i}^{}$, $Q_{{1\rm{R}}i}^{}$, $Q_{{2\rm{F}}i}^{}$ were calculated.
We selected the bending angle of fingers and the distance between the thumb and other fingers as the two evaluation index, which are the key indexes for hand grasping. The results are respectively presented 
presented in Fig. \ref{IROSFigBiomimetic}(a) and (b). The time scale defined by frame of the human and robot hands were normalized. 

At Stage 2,  the efficacy coefficients of both human and robot hand generally increase.
At Stage 4, the soft robotic hand implements reverse-direction grasping. Thus, the  $Q_{{2}i}^{}$ of the human hand during this stage is constant. Meanwhile, the  efficacy coefficients $Q_{{1\rm{R}}i}^{}$ starts to increase, which also leads the total efficacy coefficients $Q_{{1}i}^{}$ continues to increase. The trade at Stage 6 is  inverse compared with Stage 2.

For single-direction grasping, the efficacy coefficient of human hand (black line) presents a quite similar variation with that of robot hand  through all the 3 stages. This testify that the soft hand performs a human-like motion of hand grasp. When considering bidirectional grasping, the variation of $Q_{{1\rm{F}}i}^{}$ at Stage 2 is quite similar with that of $Q_{{1\rm{R}}i}^{}$. This indicates that the two successful grasps of  bidirectional grasping is mutually independent.

\section{Conclusion}

In this paper, we present a novel 3D reconstruction approach  to analyze the complex grasping motions of soft robots in underwater environment through experiment videos. Using the videos captured by the 3D imaging system, we reconstruct the configuration of the hand.  The proposed reconstruction method perform the reconstructed hand pose when the hand is both with and without grasp objects. The 3D reconstructed pose result presents a satisfactory performance and  good  accuracy. Using efficacy coefficient method, the finger contribution of the bending angle and distance between fingers of  soft robot hand is analyzed by compared with that of human hand. The results show that the soft robot hand perform a human-like motion  in both single-direction and bidirectional grasping.

\bibliographystyle{IEEEtran}
\bibliography{ref}


\end{document}